\pdfoutput=1
\documentclass[11pt]{article}

\usepackage{acl}
\usepackage{times}
\usepackage{latexsym}

\usepackage[T1]{fontenc}
\usepackage[utf8]{inputenc}

\usepackage{microtype}

\usepackage{amsmath}
\usepackage{amsfonts}
\usepackage{adjustbox}
\usepackage{booktabs}
\usepackage{graphicx}
\usepackage{subcaption}
\usepackage{caption}
\usepackage{pythonhighlight}
\usepackage{wasysym}
\usepackage{algorithm}
\usepackage{algpseudocode}

\newcommand{\algname}{\emph{SepLL} }
\title{SepLL: Separating Latent Class Labels from Weak Supervision Noise}

\author{Andreas Stephan$^{1,2}$ \\
 \\
\\ \And
  Vasiliki Kougia$^{1,2}$ \\%$^\circ$$^\star$ \\
$^1$\emph{Research Group Data Mining and Machine Learning},\\\emph{Faculty of Computer Science, University of Vienna}, Vienna, Austria \\
$^2$\emph{UniVie Doctoral School Computer Science}, Vienna, Austria \\
$^3$\emph{Faculty of Philological and Cultural Studies, University of Vienna}, Vienna, Austria \\
  \texttt{\{andreas.stephan,vasiliki.kougia,benjamin.roth\}@univie.ac.at} \\ \And
  Benjamin Roth$^{1,3}$\\%$^\circ$$^\#$ \\
  }

\begin{document}
\maketitle

\begin{abstract}
In the weakly supervised learning paradigm, \emph{labeling functions} automatically assign heuristic, often noisy, labels to data samples.
In this work, we provide a method for learning from weak labels by separating two types of complementary information associated with the labeling functions: 
information related to the target label and information specific to one labeling function only.
Both types of information are reflected to different degrees by all labeled instances.
In contrast to previous works that aimed at correcting or removing wrongly labeled instances, we learn a branched deep model that uses all data as-is, but splits the labeling function information in the latent space.
Specifically, we propose the end-to-end model \emph{SepLL} which extends a transformer classifier by introducing a latent space for labeling function specific and task-specific information.
The learning signal is only given by the labeling functions matches, no pre-processing or label model is required for our method. 
Notably, the task prediction is made from the latent layer without any direct task signal.
Experiments on Wrench text classification tasks show that our model is competitive with the state-of-the-art, and yields a new best average performance. 
\end{abstract}

\section{Introduction}

%LUISAs hints:
%Also ich finde immer Ziel der Intro ist: Leser abholen vom aktuellen Stand (hast du super gemacht), Leser mitnehmen in die neue Idee (hast du auch gut gemacht), Leser die Basics dafür an die Hand geben, wenn sie keine Experten sind (da geht noch bisschen was), Leser einen Ausblick geben auf das was kommt (hast du gemacht) und einen guten Hinweis auf die Ergebnisse (fehlt mir noch). Zu guter Letzt dem Leser sagen, warum man ne geile Sau ist (contributions) (hast du gemacht �� )
%Also nach dem Intro will ich wissen: Was hat der warum und wie im groben gemacht und was kam da raus? Dann kann ich entscheiden als Leser, ob ich den Rest lesen will und bin happy.

% Motivate why the problem you tackle is relevant
In recent years, large language modelling approaches have proven their applicability to a wide range of tasks, mainly due to the pre-training and fine-tuning paradigm.
%TODO: possibly get rid of these lines
This has created a need for large labeled datasets, as training on these datasets enables models to achieve state-of-the-art performance. 
However, obtaining manually created labels is expensive, tedious and often requires expert knowledge. 
As a consequence, significant areas of research are devoted to addressing this challenge by minimizing the need for labeled data. For example, research directions include transfer learning \cite{ruder-etal-2019-transfer} or few-shot learning \cite{brown2020language}.
Another research direction to address this challenge is weakly supervised learning. 
The idea is to use human intuitions, heuristics and existing resources, e.g., related databases, to create weak (noisy) labels.
%Importantly, in \cite{Ratner2016} the authors introduce the data programming paradigm, which provides a formalism to write so-called labeling functions (LFs), i.e. functions which create weak labels programatically.

%\begin{figure}
%\includegraphics[scale=0.4]{files/intuition_graphic.png}
%\caption{This figure represents the intuition motivating this work. On the left are text samples, in the middle are labeling %function (LF) matches and on the right it is shown how much information the sample provides for a target task.} 
%TODO: Target task vs target label, etc.
%\label{fig:intuition}
%\end{figure}

% Characterize current approaches that tackle the same problem and highlight their shortcomings (as related to your approach)
Several approaches have been proposed to increase the quality of the resulting labels. 
For example, \citet{DBLP:journals/corr/abs-1711-10160} use generative modeling to learn a probability distribution over the labeling function matches, i.e., weak labels, and unknown true labels in order to denoise the labels and subsequently train a classifier.
Recently, several works use student-teacher schemes that use knowledge inherent to pre-trained models \citep{karamanolakis2021selftraining,cachay2021endtoend,2020}.
%TODO denoise, knowman:
Usually a summary statistic of weak labels, such as majority vote, is used as ground truth and iteratively updated during training, for example by employing a regularization based on the prediction confidence of the model \citep{yu2021finetuning}. 
Thus, most methods share the property that the weak labels, i.e., the learning signals, are transformed or updated throughout the learning process. 
%TODO: rewrite that we loose multiple LF matches

%We think this diverges from the real goal. Weak supervision should aim to separate the information relevant to the task to the unrelevant information.  
%We think an idealized WS method should really separate the wanted information from the weak information and not rely on some intermediate representation.

%Currently methods try to extract the knowledge of pre-trained LLM's to learn samples where the input-output (summary statistic, e.g. majority vote) relation makes most sense together by probing the likelihood.

%While these research directions provide a path forward to tackle Weak Supervision, we think another important question is how to model and identify or disentangle the intuition provided by labeling functions which is so easy to understand for humans.

%In the long run, weak supervision also might support research to deal with 

% Describe the main novelty of your approach
Instead of updating the weak labels, we want to keep them as-is and make use of a different intuition. 
Each labeling function provides information relevant to the prediction task but also information only related to the function itself. 
%See figure \ref{fig:intuition}
Our idea is to view these two types of information as complementary and build a model which separates them.
%The idea is to view these two types of information as complementary and build a model which separates them.

\begin{figure*}[t]
\centering
\includegraphics[scale=0.6]{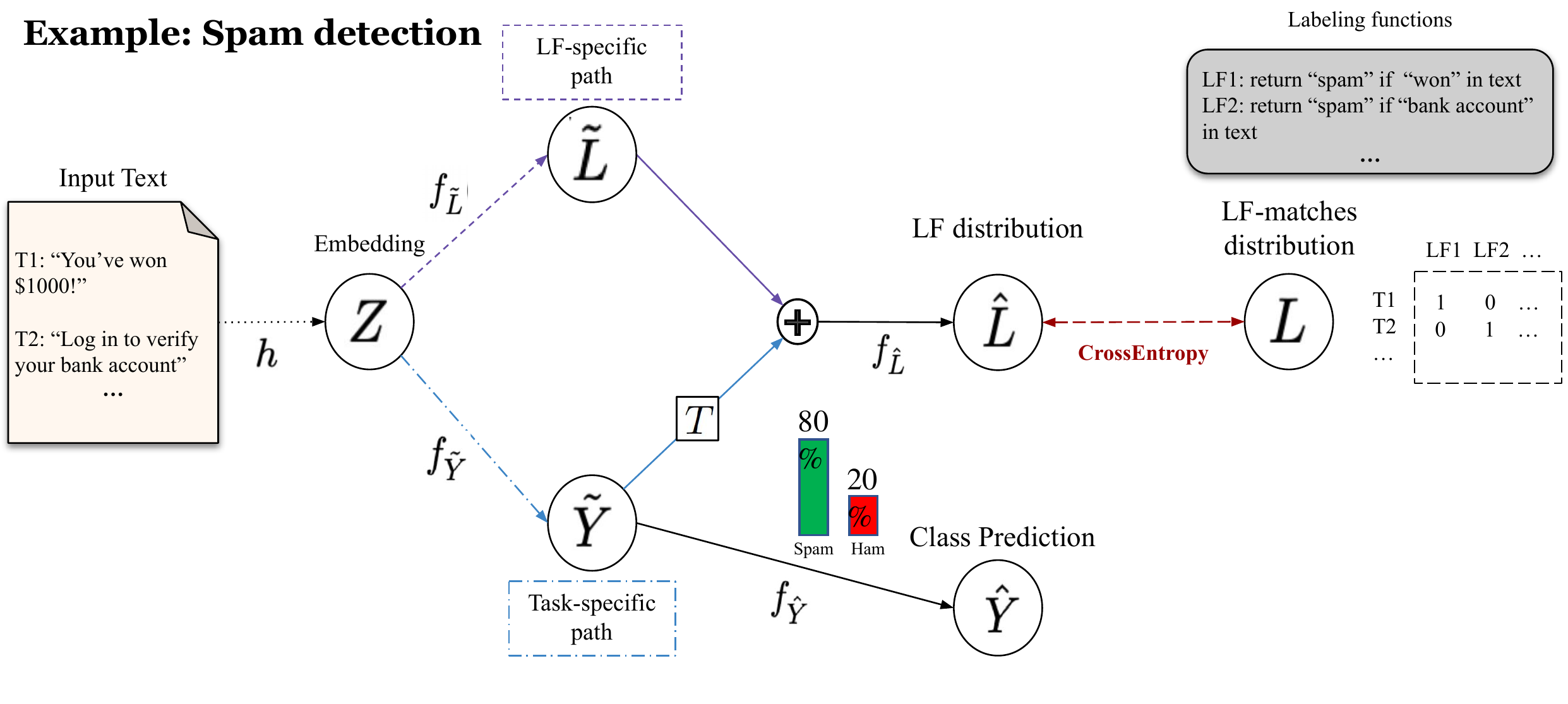}
\caption{Overview of \emph{SepLL}. Text gets embedded into $Z$ by a Transformer encoder, and then this representation is split into labeling function-specific and task-specific information.
The task-specific information is translated back into the LF space and re-combined into $\hat{L}$. %TODO:(Ben) explicitely mention T-matrix
A cross-entropy loss between the distribution of labeling function matches $L$ and $\hat{L}$ is minimized.
The latent task prediction $\hat{Y}$ can be used for classification.
}
\label{fig:model}
\end{figure*}

% Describe your methodology
To this end, we propose \emph{SepLL}, an end-to-end model that stacks two branched latent layers, representing target-task-related and labeling-function-related information, on top of a transformer encoder and recombines them for predicting labeling function occurrences (Figure \ref{fig:model}). 
Then, the learning signal is only given by the weak labels.
Notably, the task prediction is performed from the latent space without any direct supervision.
Multiple information routing strategies are employed to improve the separation.
%TODO:(Ben) mention "backtranslation" from task-space to LF-space via explicit mapping (T-matrix)?

%Based on these intuitions, we derive a simple generative structure of the prediction process and derive a model which separates "latent" knowledge (class information; semantic) and labeling function specific information (syntactic) by using a simple addition scheme and a information-theoretically motivated regularization mechanism.

% Quantify the improvement
In order to evaluate the performance, experiments on the text classification tasks of the Wrench benchmark \cite{zhang2021wrench} are performed. 
Our model achieves state-of-the-art performance when compared to standalone models as well as when combined and compared with the self-improvement method Cosine \cite{yu2021finetuning}.
%As the classification head of our model is easily extractable, we can view our model as label model and use it for the iterative end model framework Cosine. The resulting combination achieves a new state-of-the-art.
An ablation study shows the importance of each information routing strategy. The experiments show that in addition to its task performance, the model is able to memorize the labeling function information.

%TODO: Write about ablations; benefits of still having LF info

%While labeling function prediction is usually close to 100\%, it provides an interesting insight into the impact of class specific information relevant to predict a labeling function.
% Add here: how good we can understand the LF info based on what type of signal

%As the classificator of our model is easily extractable, we can view our model as end model, but it is also a valid label model, allowing it to be plugged in much more complicated end models such as COSINE.  Our experiments show results competitive with state-of-the-art.

% Highlight the main insight
The contributions can be summarized in three parts:
1) We introduce a new intuition about the information provided by labeling functions and turn it into a method, \emph{SepLL}, reflecting the intuition in the latent space.
2) We provide an analysis through experiments on the Wrench benchmark, an ablation study and an in depth analysis of the two latent spaces.
3) We provide the code and a suitably transformed version of the input data. \footnote{https://github.com/AndSt/sepll}
\section{Related Work}

\textbf{Weak Supervision.} A main concern in machine learning is that a large amount of labeled data is needed in order to train models that achieve state-of-the-art performance. Among others, the field of weak supervision aims to address this issue. The idea is to formalize human knowledge or intuitions into weak supervision sources, called labeling functions, which can be used to produce weak labels. Examples of labeling functions are heuristic rules, e.g., keywords, regular expressions, other pre-trained classifiers or knowledge bases in distant supervision \cite{craven1999constructing,mintz-etal-2009-distant,hoffmann2011knowledge,takamatsu-etal-2012-reducing}.
%Weak labels produced by the multiple labeling functions have to be aggregated (label model) and can then be used to train an end model (end model). 

%\citet{DBLP:journals/corr/abs-1711-10160} proposed Snorkel, which uses the Data Programming paradigm \cite{Ratner2016} to train machine learning models with weak labels. 

%\cite{Ratner2016,Ratner2019,2020}, but a one-stage approach has also been employed \cite{2020,karamanolakis2021selftraining,Lan2020}. 
%Even though the weak labels are easier and cheaper to obtain than human annotated labels, they are also noisy and of lower quality. 

A main challenge that appears in a weak supervision setting is how to create accurate labeling functions and how to unify and denoise them. Majority vote, Snorkel \citep{DBLP:journals/corr/abs-1711-10160} (based on data programming)
and Flying Squid \citep{DBLP:conf/icml/FuCSHFR20}
are methods that compute weak labels based on generative models over the labeling function matches and unknown true labels.
These models are referred to as label models.
Subsequently so called end-models, e.g., BERT-style classifiers \cite{devlin2019bert}, or methods dedicated to noisy training labels are used to train a final model.

Recently, neural methods, including the use of pre-trained models, gained more traction.
%and the use of pre-trained models was introduced in order to add more knowledge and achieve better generalization. 
\citet{cachay2021endtoend} use a classifier and a probabilistic encoder for the labeling function matches and optimize them using a noise-aware loss. Similarly, \citet{2020} combine a classifier and a attention-based denoiser, but also include unlabeled samples.
%First, Snorkel provides to users an interface to write labeling functions and then performs denoising by learning a generative model over the labeling functions. 
%The final output of Snorkel are probabilistic weak labels that can be used to train machine learning models.
\citet{yu2021finetuning} introduced Cosine, which is a method to self-optimize classification models.
They leverage contrastive learning and confidence regularization, i.e., high-confidence samples, to optimize a model's performance.

Other approaches use additional signals. For instance, ImplyLoss \cite{awasthi2020learning} uses access to exemplars, i.e., single, correctly labeled samples and
ASTRA \citep{karamanolakis2021selftraining} follows an attention based student-teacher mechanism with an additional supervision of a few manually annotated labeled samples.
%It employs rules, unlabeled, and a few manually annotated data. The teacher, which is a Rule Attention Network (RAN) is used to aggregate the weak labels obtained from rules and pseudo-labels produced by the student. 
%The aggregated labels are used to re-train the student, thus following a iterative teacher-student training.
\citet{zhu2022meta} uses a meta self-refinement approach which makes use of access to the validation performance.

Our experiments are built on the Weak Supervision Benchmark (Wrench) \citep{zhang2021wrench}, which is a framework that aims to provide a unified and standardized way to run and evaluate weak supervision approaches. 
A wide range of tasks, datasets and implementations of weak supervision methods are available.
%They perform extensive experiments to compare these methods and their variants, and they continuously update the framework with new datasets and methods.

\textbf{Latent Variable Modelling.} Existing work regarding latent variable modelling in different areas of machine learning has influenced the rationale behind this work. 
Research in representation learning has focused on modelling mutually independent factors of variation, e.g., color in computer vision, explicitly in some latent space. 
Often this is called \textit{disentanglement} \cite{article}. 
This is transferable to our setting as we aim to obtain the task prediction as a disentangled factor.
An important early technique is Independent Component Analysis (ICA) \cite{COMON1994287}.
%TODO: one sentence
\citet{kingma2014autoencoding} introduced variational autoencoders (VAE's) to neural networks, allowing complex data distributions to be represented as simple distributions in the latent space.
%TODO better
An extension is given by $\beta$-VAE \cite{Higgins2017betaVAELB}, which is more suitable for disentanglement.
% theoretical latent stuff
In addition, there has been progress on theoretical work, which aims to give an insight on what information is identifiable by using self-supervised learning (SSL), e.g., \citet{Zimmermann2021ContrastiveLI} prove under certain assumptions that it inverts the data generation process. 
An interesting perspective is the separation of content and style, e.g., the animal in a picture (content) and the camera angle of the image (style). Under milder assumptions as in \citet{Zimmermann2021ContrastiveLI}, it is proved by \citet{vonkuegelgen2021selfsupervised} that this separation is achieved using SSL.
Mentioned works are not directly applicable to our task, because we want to separate general aspects of labeling functions, which are useful for prediction tasks, from labeling function specific aspects.
Another line of research models the true distribution in the latent layer \citep{c51d68a3106242f08ed001d0c46320b3,Goldberger2017, Bekker2016} while training on the noisy training labels. The typical assumption is that the noise distribution only depends on the class. In weak supervision the noise depends on the input, by definition of labeling functions, thus these types of assumptions are not directly applicable.

%All either ask for multiple quantities, which are clearly separable from a human point of  view, or are vector representations, which are not directly interpretable by humans.
%This differs from our work as we want to understand an interpretable quantity (prediction task), while at the same time not losing the rest of the information (labeling function specific information) which is not explicitly interpretable by humans.

% New references
%Another related line of research aims to tackle the problem of noisy labels with latent variable modelling. 
%\citet{Xiao2015} proposed a probabilistic model that corrects noisy image labels by treating true labels and noise types as latent variables and learning the relations between them and the noisy labels \citep{Xiao2015}.
%\citet{Sukhbaatar2014} introduced a linear noise layer on top of the softmax layer of a Convnet classifier and train it end-to-end in order to learn the noise distribution.
%Similar to the approaches mentioned above, \citet{Bekker2016} proposed a probabilistic network to model the noise in labels. 

%However, unlike previous works, they do not use any clean labeled data. Similar to our work, \citet{Goldberger2017} treat the true labels as a latent variable. But they make the assumption that the noise is only dependent on the class variable which is not applicable in weak supervision.

\section{Method}
\label{sec:method}

The motivation of this work is that each labeling function provides two types of information. % which are separable. 
On the one hand, it provides information about the target task, e.g., spam detection, and on the other hand it provides information related to the labeling function itself.
This translates to our model, called \emph{SepLL}, which aims to separate these two types of signals in a latent space. Figure \ref{fig:model} provides an overview of \emph{SepLL}.

In this section, we first introduce some notation and then describe the architecture of \emph{SepLL}. Following, the training mechanisms, which aim to support the separation of the two information types are discussed.

% The main goal of this section is to discuss the basic architecture, how it is related to the described intuition, and how we steer the learning process.

% %how we steer build the architecture and how we steer the learning process in order to make a prediction from the latent space. 
% After the introduction of some notation, the basic model is discussed.
% Afterwards, we provide training details which aim to support the separation of the two information types. More specifically regularization, noise injection and smoothing by the usage of unlabeled samples, i.e. labels where no weak label matches, is discussed. 

\subsection{Problem Setup and Notation}

In general, the goal is to solve classification tasks, e.g., spam detection asks whether a text is spam or not.
The input space is denoted by $X$ and the unknown labels are denoted by $Y=\{y_1, \dots ,y_c\}$. Additionally $m$ labeling functions $l_i:X \rightarrow \{ y \} \cup \emptyset, i=1, \dots, m$ are given where each labeling function (LF) either assigns a dedicated specific label $y \in Y$ to a sample or abstains from labeling. 
If a label is assigned, we say a labeling function \textit{matches} a sample.
The task is to use input $X$ and labeling functions $l_i$ to learn a mapping $X \rightarrow Y$.
We use the format of the Knodle \cite{sedova-etal-2021-knodle} framework, where each labeling function is encoded as a labeler for exactly one class.
This is in contrast to other conventions where a single labeling function is allowed to label multiple classes, e.g., in \citet{Ratner2016}.
This convention can easily be transformed into our setting, by splitting multi-class LFs into multiple class-specific LFs.
The matching matrix $L \in  \{0, 1\}^{n \times m}$ describes whether labeling function $j$ matches sample $i$ by setting $L_{ij}=1$, otherwise $L_{ij}=0$. The mapping matrix $T \in \{0, 1\}^{m \times |Y|}$ reflects a simple mapping between labeling function $i$ and class $j$ by $T_{ij}=1$, otherwise $T_{ij}=0$. 
%TODO: write that?
%This injective model enables the definition of our model.

\subsection{Basic Model}

First, the model transforms input text $X$ into a latent representation $Z$ using an encoder $h: X \rightarrow \mathbb{R}^d$. 
In this case, a pre-trained transformer encoder transforms input text $x \in X$ into the <CLS>-token embedding $z=h(x) \in \mathbb{R}^d$.

Following, there are two transformations $ f_{\Tilde{Y}}:\mathbb{R}^d \rightarrow \mathbb{R}^{|Y|}$ and $ f_{\Tilde{L}}: \mathbb{R}^d \rightarrow \mathbb{R}^m$, 
which are realized by two multi-layer perceptrons.
%Andy: Is the next sentence is reasonable?
The goal is to train the model such that $ f_{\Tilde{Y}}(x)$ reflects the task information and $ f_{\Tilde{L}}(x)$ the remaining LF-related information.
Note that the resulting output represents the log-space and the transformation to probabilities happens later.

Afterwards, the two latent layers are combined again and compared to the training signal, which is purely given by the LF matches $L$.
The $T$ matrix is used to map the target label information to the corresponding LF information by
\begin{align}
\label{eq:combine_path}
f_{\hat{L}}(z) = f_{\Tilde{Y}} (z) T^{\top} + f_{\Tilde{L}} (z) \in \mathbb{R}^{m}
\end{align}
where $z=h(x)$ is the latent representation.
Crucially, the $T$ matrix establishes the connection between task path and LF path.
%TODO: Discussion of importance of T matrix

In order to run the optimization we compare the combined signals to the labeling functions matches. 
Therefore we define the \textit{LF distribution} as the normalized $L$ matrix, i.e., $P_{ij} = \frac{L_{ij}}{\sum_{k} L_{ik}}$.

%Alternatively one could see every entry as a random variable and make multiple predictions. We realized that hurts performance, ....

Finally, the loss is computed as the cross entropy between the labeling function distribution and the prediction

\begin{align}
  \text{CE}(P, Q) = \frac{1}{n} \sum_{i=1}^n \sum_{j=1}^m P_{ij} \log( Q_{ij} )
\end{align}
where the prediction probability $Q$ is given by a softmax activation over $f_{\hat{L}}$. %TODO explain this in more detail?
%is given by $p=\text{softmax}(f_{\hat{L}}(z))$

The task prediction is computed using a softmax activation on the latent task signal, i.e.,
\begin{align}
    P(y_i | x) = \left(f_{\hat{Y}}(z)\right)_i = \frac{e^{ \left(f_{\Tilde{Y}}(z)\right)_{i}}}{\sum_{j=1}^c e^{\left(f_{\Tilde{Y}}(z)\right)_j}}
\end{align}
where $z=h(x)$ is again the latent representation. Thus, no direct supervision is performed.

\subsection{Latent information routing}  

%TODO: Other titles: latent information updating, gradient update flow, etc.
Clearly, the separation could easily collapse by just using the labeling function path, i.e., there is no apparent reason why the LF prediction $\hat{L}$ is not only based on the LF-specific path. 
%As our results show, there is a need to steer the learning process. 
Therefore we introduce three schemes supporting the separation of information.
Firstly a regularization, secondly an adaption of the learning label and thirdly the inclusion of unlabeled samples are discussed.

%based on ideas from standard optimization, how the two flows %TODO what are they
%combined and from the weak supervision literature

\textbf{Regularization.}
The easiest solution to the problem is to introduce standard regularization schemes. 
We consider the $L_2$ regularization as suitable because it encourages the optimization to put similar weight on the parameters.
We test two types of $L_2$ regularization. 
Firstly, standard weight decay is used, which employs an $L_2$ regularization on all parameters, including the transformer encoder.
Secondly, an additional $L_2$ regularization is applied to $f_{\Tilde{L}}$.
%TODO redo next sentence
The goal is to regularize the labeling function path such that more weight is put on the class prediction path.
In the experimental part we refer to the two types as weight decay and $L_2$ regularization, respectively.

\textbf{Noise Injection.} 
Our hypothesis is that the optimization routine puts weight on the task-specific path if two labeling functions belonging to the same class match simultaneously.
But most of the time samples are only matched by a single labeling function.
Thus, we inject noise in the form of additional \emph{``hallucinated''} matches into the labeling function matrix $L$.
If a sample is matched by a labeling function, we create a match for all other labeling functions for the same class with a probability proportional to a random factor $\lambda \in [0, 1]$, which is a hyperparameter.
%TODO: thus we just adapt training signal Q, ...

%TODO: just by binomial ...
%Assume sample $x_i$ has only a single match, e.g. $L_{ij}=1, L_{ik}=0$ for all $k \neq j$. Then, if if labeling function $k$ is a labeler for the same class as function $j$, i.e. $T_{js}=T_{ks}=1$, we set $L_{ik}=1$ with probability $r$ which is a hyperparamater.
%TODO: thus we just adapt training signal Q, ...

\begin{table*}[ht]
\centering
\begin{tabular}{lrrrrrr}
\toprule
 Dataset &  \#Classes &  \#LF's &  \# Train & \# Coverage &  \# Dev &  \# Test \\
\midrule
    IMDb &         2 &      9 &    20000 &            88\% &   2500 &    2500 \\
    Yelp &         2 &     15 &    30400 &            83\% &   3800 &    3800 \\
 Youtube &         2 &     10 &     1586 &            88\% &    120 &     250 \\
     SMS &         2 &     73 &     4571 &            41\% &    500 &     500 \\
  AGNews &         4 &      9 &    96000 &            69\% &  12000 &   12000 \\
    TREC &         6 &     68 &     4965 &            95\% &    500 &     500 \\
  Spouse &         2 &      9 &    22254 &            26\% &   2811 &    2701 \\
 SemEval &         9 &    164 &     1749 &           100\% &    178 &     600 \\
\bottomrule
\end{tabular}
\caption{Statistics describing the datasets as they are used in the WRENCH framework. Coverage is computed on the train set by  dividing the number of samples having at least one match by the number of samples. }
\label{table:dataset_stats}
\end{table*}

\textbf{Usage of Unlabeled Samples.}
It has been previously shown, e.g., by \citet{2020} and \citet{yu2021finetuning} that it is effective in weak supervision settings to make use of unlabeled samples $X_U$, i.e., samples where no labeling function matches. Apart from a performance increase it was shown that the learning gets more robust.
Thus, we adapt this semi-supervised learning approach.
In order to comply with our basic model, we need to create a labeling function distribution $P_U$.
We take the simplest idea and define $P_U$ as the uniform distribution over the labeling function matches, i.e., $(P_U)_{ij} = \frac{1}{m}$ for all unlabeled samples $i$ and labeling functions $j$.

\section{Experimental Setup}

Before analysing the experiments, we discuss the experimental setup. 
More specifically, an overview of the used datasets, baselines, hyperparameters and some notes on reproducibility and implementational details are given.

\subsection{Datasets} 

For the experiments we used eight text classification datasets that are currently included in the Wrench benchmark. As shown in Table \ref{table:dataset_stats} the datasets reflect varying properties, such as sample size, coverage or the amount of labeling functions.

Five out of eight datasets represent binary classification problems. These are: (i) \textbf{Youtube \citep{7424299}}, which consists of text comments from YouTube videos, each labeled as spam or non-spam, (ii) \textbf{SMS \citep{Almeida2011ContributionsTT}}, which is a mobile phone spam corpus, which contains real SMS messages that are spam or non-spam, (iii) \textbf{IMDb review dataset \citep{maas-etal-2011-learning}} contains reviews from IMDb and each review is labeled as positive or negative, (iv) \textbf{Yelp \citep{Zhang2015}} consists of positive or negative reviews from the Yelp Dataset Challenge 2015, (v) \textbf{Spouse \citep{Corney2016WhatDA} }, which is a relation classification dataset, where we decide for each sentence if it contains a spouse relation or not. Three datasets correspond to multi-class problems: (i) \textbf{AGNews \citep{Zhang2015}} consists of news articles classified into 4 classes, (ii) \textbf{TREC \citep{Li2002}} is a question classification dataset, where the questions are classified into 6 labels, %TODO: maybe which clsses
 and (iii) \textbf{SemEval \citep{Hendrickx2010semeval}} contains sentences collected from the Web and the task is to identify the relation between two nominals tagged in each sentence among 9 types of semantic relations.

\begin{table*}[h]
\small
    \centering
    \begin{tabular}{lrrrrrrrrr}
\toprule
            &  IMDb &  Yelp &  Youtube &   SMS &  AGNews &  Trec & Spouse &  Semeval &      Avg. \\
\midrule
    \textbf{Supervised}    & & & & & & & & & \\
     Gold+RoBERTa$^\dagger$ & 93.25 & 97.13 &    95.68 & 96.31 &   91.39 & 96.68 &      - &    93.23 & 88.58 \\
               \hline
               \textbf{Statistical}    & & & & & & & & & \\
               MV$^\dagger$ & 71.04 & 70.21 &    84.00 & 23.97 &   63.84 & 60.80 &  20.81 &    77.33 & 59.00 \\
               DP$^\dagger$ \cite{DBLP:journals/corr/abs-1711-10160} & 70.96 & 69.37 &    82.00 & 23.78 &   63.90 & 64.20 &  21.12 &    71.00 & 58.29 \\
               FS$^\dagger$ \cite{DBLP:conf/icml/FuCSHFR20} & 70.36 & 68.68 &    76.80 &  0.00 &   60.98 & 31.40 &   34.3 &    31.83 & 46.79 \\
               \hline
               \textbf{Neural}    & & & & & & & & & \\
          WeaSEL \cite{cachay2021endtoend} & 85.16 & 91.23 & 96.40 & 2.94 & 85.92 & 64.2 & 0.00 & 44.30 & 58.77 \\
          Denoise$^\dagger$  \cite{2020} & 76.22 & 71.56 &    76.56 & 91.69 &   83.45 & 56.20 &  22.47 &    80.83 & 69.87 \\
          KnowMAN \cite{marz-etal-2021-knowman} & 59.00 & 76.76 & 94.00 & 92.80 & 84.68 & 65.20 & 25.48 & 80.50 & 72.30 \\
       RoBERTaMV$^\dagger$ & 85.76 & 89.91 &    96.56 & 94.17 &   \textbf{86.88} & 66.28 &  17.99 &    84.00 & 77.69 \\
       RoBERTaDP$^\dagger$ & 86.26 & 89.59 &    95.60 & 28.25 &   86.81 & 72.12 &  17.62 &    70.57 & 68.35 \\
        RoBERTaFS$^\dagger$ & \textbf{86.95}  & \textbf{92.08} &    93.84 & 10.72 &   86.69 & 30.44 &    0.0 &    31.83 & 54.07 \\
        \hline
     \algname & 83.57 & 91.32 &    \textbf{97.50} & \textbf{95.45} &   85.47 & \textbf{81.25} &   \textbf{43.2} &    \textbf{87.33} & \textbf{83.14} \\
     \hline
     & & & & & & & & & \\
     \hline
     \textbf{+Cosine} \cite{yu2021finetuning}    & & & & & & & & & \\
     RobertaMV+Cosine$^\dagger$ & \textbf{88.22} & 94.23 &    \textbf{97.60} & 96.67 &   \textbf{88.15} & 77.96 &   40.5 &    86.20 & 83.69 \\
 RobertaDP+Cosine$^\dagger$ & 87.91 & 94.09 &    96.80 & 31.71 &   87.53 & 82.36 &  28.86 &    75.17 & 73.05 \\
% FS+Cosine & 87.65 & 95.45 &    95.20 & 82.24 &   87.73 & 38.80 &  16.06 &    31.83 & 66.87 \\
%& 88.00 & 95.07 &    97.60 & 97.01 &   86.28 & 83.40 &  43.37 &    86.83 & 84.70
  \emph{SepLL}+Cosine & 88.00 & \textbf{95.07} &    \textbf{97.60} & \textbf{97.01}&   86.28 & \textbf{83.40} &  \textbf{43.37} &  \textbf{86.83} & \textbf{84.70} \\
\bottomrule
\end{tabular}
    \caption{Results on the Wrench benchmark tasks. Accuracy values are reported for all datasets, except for SMS and Spouse where the binary $F_1$ is shown due to class imbalance. Numbers directly taken from the Wrench paper are marked by $\dagger$.}
    \label{tab:results}
\end{table*}

\subsection{Baselines} 

We compare \algname 
to several traditional and state-of-the-art models that can be categorized in 4 approach types: supervised, statistical, neural and cosine based (see Table \ref{tab:results}). 
Wherever possible the RoBERTa-base \cite{liu2020roberta} backbone model is used.
%TODO Weasel

For the supervised approach (\textbf{Gold + RoBERTa}), which serves as an upper bound, we perform a standard fine-tuning of a RoBERTa \cite{liu2020roberta} pre-trained model using gold labels. 
% Except for the Spouse dataset, labeled data is available. \\

The statistical approaches are: (i) \textbf{Majority Vote (MV)} As the name suggests, majority vote picks the label indicated by the majority of labeling function matches. In case there is no match, a random label is chosen, (ii) \textbf{Data Programming (DP)} employs Snorkel DP \cite{DBLP:journals/corr/abs-1711-10160} to obtain weak labels and (iii) \textbf{Flying Squid (FS)} \cite{DBLP:conf/icml/FuCSHFR20} is a label model making use of so-called triplet methods.

In the neural category of baselines there are three dedicated end-to-end models: (i) \textbf{WeaSEL} \cite{cachay2021endtoend} uses a classifier and a probabilistic encoder combined with a noise-aware loss function, (ii) \textbf{Denoise} \cite{2020} uses a classifier and an attention-based denoiser, mutually optimizing each other, (iii) \textbf{KnowMAN} is an adversarial architecture that aims to learn representations that are invariant to the labeling function signals \cite{marz-etal-2021-knowman}. 
In addition to learning a classifier for the end task, a labeling function discriminator is trained and its negative gradient is used to update a shared feature extractor.
%KnowMAN is able to control with a hyper-parameter to what degree the labeling function-specific information should be suppressed. 
%&To be comparable to the other baselines, KnowMAN was trained based on RoBERTa for this paper. \\

We also employ models that combine the labels obtained by MV, DP and FS with standard supervised learning using the \textbf{RoBERTa} model, which we call \textbf{RoBERTa-MV}, \textbf{RoBERTa-DP} and \textbf{RoBERTa-FS} respectively \cite{liu2020roberta}. 

\textbf{Cosine} \cite{yu2021finetuning} takes a pre-trained classifier and uses contrastive learning and confidence regularization to improve the  performance of the classifier.
Cosine is a model-agnostic method for self-training that can be combined with any classifier and is not specific to weak supervision.
It has been observed that Cosine particularly helps with standard weak supervision methods like majority voting and data programming, and we include the best-performing combinations from Wrench (\textbf{RoBERTa-MV+Cosine}, \textbf{RoBERTa-DP+Cosine}), as well as \emph{SepLL}+Cosine in our experiments.
Additional information on the baselines is given in appendix \ref{appendix:experimental_description}.

\subsection{Hyperparameters } 

Two types of hyperparameters are tuned, namely transformer related hyperparameters and information routing related ones. 
We perform a grid search on these and take the best model based on the validation set. 
The first group includes a learning rate in $\{$1e-5, 2e-5, 5e-5$\}$, a batch size of $16$ and warmup steps in $\{0, 100\}$.
For information routing related hyperparameters we search for weight decay in $\{0, 0.01, 0.001\}$, $L_2$-regularization on the LF path in $\{0.1, 0.5, 0. , 1.  \}$, and for noise injection  $ \lambda \in \{0.  , 0.1 , 0.2 , 0.05 \}$.
\algname is always trained using AdamW \citep{loshchilov2018decoupled}.

\subsection{Implementation and Reproducibility} 

Our implementation is in JAX \cite{jax2018github}, using the Huggingface transformers library \cite{wolf2020huggingfaces} as a high level interface.
%TODO: What about next line:
%When we used models from the WRENCH framework, which are based on pre-trained models, we tak JAX checkpoints are transformed to PyTorch \cite{falcon2019pytorch} and are used as backbone in WRENCH.

%Most of the numbers are taken directly from the WRENCH paper. 
%We found this beneficial as this saves resources and ensures a common standard. 

In order to save resources, we report numbers directly from the Wrench paper, whenever applicable.
In other cases, we use the datasets and the evaluation setting of Wrench, and the original implementations and reported hyper-parameters of the respective publications.

We use RoBERTa-base as backbone of most models, so all models use approximately 125 million parameters. Fine-tuning one instance takes between 5 and 60 minutes, depending on the dataset size. The experiments are performed on a Nvidia DGX-1 using a single Tesla V100 graphics card per run.
%To be concise, we reproduce majority vote on all datasets and the combination of majority vote and Roberta on IMDb in JAX. Then we transform the latter to PyTorch and use the WRENCH framework to reproduce the combination of MV, Roberta and Cosine.

%In order to ensure comparability with our JAX implementation we reproduce selected methods. %TODO:(Ben)???
%To be concise, we reproduce majority vote on all datasets and the combination of majority vote and Roberta on IMDb in JAX. 
%Then we transform the latter to PyTorch and use the WRENCH framework to reproduce the combination of MV, Roberta and Cosine.

\section{Experiments}

This section provides an analysis of the capabilities of the model.
After the general performance analysis, an ablation study of the information strategies is performed.
Furthermore, we investigate how much information flows through the path associated with the latent class prediction, and how much though the other, LF-specific path. % TODO:(Ben) define LF early on
%from the training signal is preserved. 
%TODO and what else is in experiments???

\subsection{General Performance}

The results are split into two parts. 
Firstly, a comparison with the standard baselines is given.
Secondly, we analyse the impact of Cosine separately, as we view it as a general framework to self-optimize the prediction performance of any pre-trained classifier.
%Thus it can work with any classifier which uses a latent representation before the classification head.
The results are shown in Table \ref{tab:results}.
%TODO: Now with flying squid also yelp, @Ben: Keep it in?
\algname outperforms the standard baselines on all tasks except IMDb and AGNews. 
We think this is due to the fact that both datasets are rather large while having a low number of labeling functions. 
%TODO: How to say that main updates should be when two simultaneously hit
The same trend is observable when combined with Cosine. 
On most tasks a new state-of-the-art is reached. 
A negative exception is given by SemEval, where Cosine is not able to generalize from our base model.
Importantly, the average performance over all tasks is improved by a margin of $6\%$ on the standard baselines and $1\%$ in the Cosine section, relative to the respective best-performing comparison method.

\begin{table}
    \centering
\begin{tabular}{lr}
\toprule
           &   avg \\
\midrule
    Full Model & 83.14 \\
   $-$ Weight decay & 82.73 \\
          $-$ $L_2$-reg. & 82.59 \\
 $-$ Unlabeled data & 82.11 \\
 $-$ Noise injection & 81.77 \\
     Basic model & 80.85 \\
\bottomrule
\end{tabular}
\caption{Ablation of information routing strategies. Each strategy is removed individually. The basic model does not use any strategy. The average is taken over all datasets on the test set.}
\label{tab:ablation_small}
\end{table}

\begin{figure*}[h]
\centering
\begin{subfigure}{.31\textwidth}
  \centering
  \includegraphics[width=\linewidth]{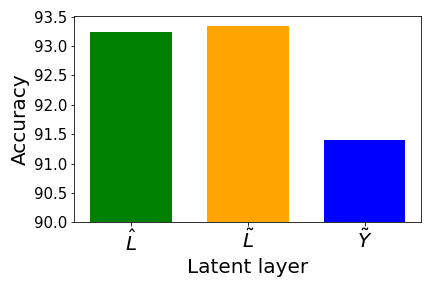}
  \label{fig:sub1}
\end{subfigure}%
$ $ $ $ $ $ $ $
\begin{subfigure}{.31\textwidth}
  \centering
  \includegraphics[width=\linewidth]{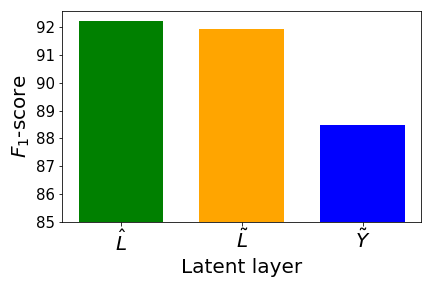}
  \label{fig:sub2}
\end{subfigure}
$ $ $ $ $ $ $ $
\begin{subfigure}{.31\textwidth}
  \centering
  \includegraphics[width=\linewidth]{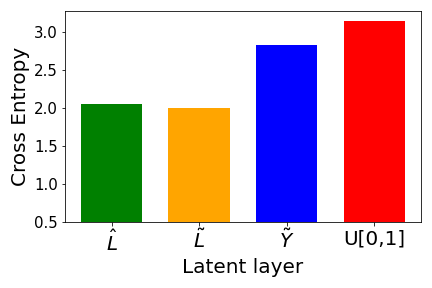}
  \label{fig:sub3}
\end{subfigure}
\caption{The figure shows the accuracy (left), the $F_1$-score (center) and the cross-entropy (right) between true labels $L$ and the predictions based on the full model $f_{\hat{L}}$ and the latent representations $f_{\Tilde{Y}}, f_{\Tilde{L}}$ predictions. On the right an additional comparison to the uniform distribution is presented.
}
\label{fig:lf_memorization}
\end{figure*}

\subsection{Ablation of Information Routing Strategies}
\label{exp_ablations}

In order to understand the impact of the routing strategies we perform an ablation study, which is shown in Table \ref{tab:ablation_small}.
%TODO: better description
Starting from the full model, each routing strategy is removed individually, and in the last row all strategies are removed.
The average performance over all datasets is shown. 
Given that there are different types of metrics this is just an aggregate view of the performance -- a detailed ablation including the performance per task is given in Appendix \ref{appendix:ablation}.

We observe that each strategy helps the performance of the end model and that the noise injection strategy has the highest positive impact. 
This reinforces the initial assumption that a sample has a larger learning impact on the task-specific path if multiple labeling functions belonging to one class match simultaneously.

Nevertheless, the basic model (without active routing) performs decent in comparison to the baselines in Table \ref{tab:results}.
%TODO: describe next sentence better
This is surprising because there is no apparent incentive to prevent the model to collapse to the LF path.
One possible explanation is that the gradient updates nevertheless flow through both paths and still update the task path in a way that it is consistent with the LF-prediction.
Moreover, predicting the LF through the task-path, albeit less accurate, is more parameter efficient than going through the LF-path.

%On the contrary it is efficient for the learning algorithm to route updates through the class path, because the optimization only needs to update one weight task prediction related weight instead of multiple labeling function weights.%TODO:(Ben) doesn't this contradict the previous sentence?

\subsection{Labeling Function Memorization}
\label{sec:lf_memorization}

As a by-product of the architecture, it is possible to predict whether a labeling function matches or not.
%TODO: Next sentence not good.
As these matches represent the learning signal, an evaluation of the prediction of labeling function matches also shows how much information is retained.
In Figure \ref{fig:lf_memorization} multiple metrics are computed to analyze the memorization of matches.  
The accuracy and the $F_1$-score (because the matrix $L$ is sparse) for LF-predictions, averaged over all test sets, are computed to measure memorization of the labeling functions matches. 

In order to transform logits into predictions, we compute the softmax activation on the functions $f_{\Tilde{L}}$ and $f_{\hat{L}}$, and define the prediction as
\begin{align*}
    L_{ij}=1 \Leftrightarrow \text{softmax}\left(f_*(z_i)\right)_j > \frac{4}{m}
\end{align*}
otherwise $L_{ij} = 0$, i.e., if a sample has $4$ times the probability of the uniform distribution. We tested the threshold $k=2,3,4$ on the dev set of IMDb and TREC and took the best performing one. Note that $k=3, 4$ performed almost equal.
To compute the task predictions, we apply the same threshold to the softmax of the mapped task logits, i.e., $\text{softmax}\left(f_{\Tilde{Y}}(z_i) T^\top \right)$.
We call the outputs of the softmax prediction distribution.
The diagram on the right in Figure \ref{fig:lf_memorization} displays the cross-entropy between the LF distribution $P$ (see section \ref{sec:method}) and the prediction distributions, and the uniform distribution $U[0,1]$.

The results show that the prediction from the final output and from the LF path are nearly identical.
Interestingly, the latent LF path achieves a slightly better accuracy, but slightly worse $F_1$-score.
In comparison to the full and the latent model, the cross-entropy between $\Tilde{Y}$ and $P$ is high where the latter approaches the cross-entropy between $\hat{L}$ and the uniform distribution.
These results indicate that $\Tilde{Y}$ does not substantially influence the prediction of $L$.
Still, the prediction made from the task-related path $f_{\Tilde{Y}}$ reaches an average $F_1$-score of $88.5\%$, which shows that it still contains much of the information needed to predict LF matches.
In appendix \ref{appendix:lf_memorization} an additional figure shows that the difference between performance on the training set and the test set is low, indicating that there is little to no overfitting towards the training LF distribution. 

\subsection{Impact of Number of LF Matches}
\label{sec:impact_lf_matches}

\begin{figure}[t]
    \centering
    \includegraphics[scale=0.5]{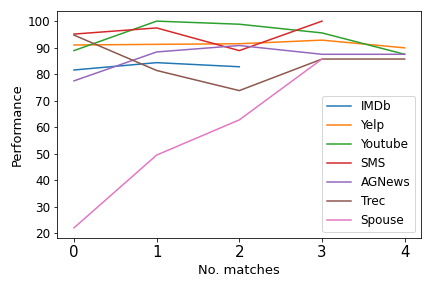}
    \caption{Performance (accuracy or F1 score) on the tasks given the amount of labeling function matches (test set).}
    \label{fig:performance_by_matches}
\end{figure}

One could expect that it is easier to predict the class label for samples with many labeling functions matches.
Figure \ref{fig:performance_by_matches} shows the performance of each task on the test set in relation to the number of labeling function matches (note that we use those LF-matches only for analysis purposes, they are not observed by the model) .
The figure does not contain SemEval as it has exactly one match per sample.

Interestingly, the performance on samples with no LF-match is, for most datasets (apart from Spouse), almost on par with samples that include patterns from one or more LFs.
It is therefore fair to say that the model generalizes beyond the labeling function information.

An exception is given by the Spouse dataset, which has the lowest coverage in the training set (see Table \ref{table:dataset_stats}).
%Thus we believe that the model overfits to the labeled samples. 
%TODO: Do we add table together or two tables?
Appendix \ref{appendix:impact_matches} provides additional numbers, including a table which shows the amount of samples per number of LF-matches per dataset.

%\section{Relation of LF prediction and task prediction}

%\subsection{Different questions}

%\begin{itemize}
%    \item Use other evaluations from E2E paper.
%    \item How to hightlight need for trafo encoder?
%    \item What happens when predicting from latent layer vs. from last layer
%\end{itemize}

%\subsection{Qualitative}

%Find examples, where multiple matches are there

%\input{chapters/discussion}

\section{Conclusion}

% Summary
This work tackles weak supervision in a novel way. 
Instead of denoising the weak labels or iteratively updating them, the weak labels are used as the training target as-is.
We introduce the \emph{SepLL} model, which separates information relevant for the target task through the usage of two latent spaces. 
One latent space is used to perform downstream task prediction, the other one aims to keep the labeling function specific information. 
Thus, the prediction is made from the latent space without any direct supervision.
Experiments show that the model is able to achieve state-of-the-art performance. %TODO
An additional investigation shows that the model is able to memorize the labeling function information. 
Hence, the model cannot only be used for downstream tasks but also to predict labeling function matches.

% % - something with separation, and the proba related to it

\section{Limitations}

As in the experiments in the Wrench benchmark, we use validation data for early stopping. 
If a setting was purely weakly supervised, with no annotated data, this would not be possible.
To ensure comparability, we stick to the Wrench setup, which assumes that the cost of annotating a small sample size for validation is acceptable in most scenarios.

Weak supervision performance highly depends on the quality and other properties of the labeling functions.
In contrast to more standardized scenarios of machine learning, e.g. in supervised learning, where it can be assumed that training and test data samples are drawn i.i.d. from the same distribution, such arguments cannot be made about the labeling functions and their relationship to the correct annotations.

Formal analysis, as it has been attempted sometimes for weak supervision, relies on heavy assumptions, which are typically unrealistic and likely to be wrong.
Such assumptions could be: the weak labelers produce noise following a defined noise distribution; the weak annotations are independent given the class label; each weak labeler exceeds a threshold accuracy; etc.
In this paper, we do not attempt a formal analysis based on such assumptions. 
However, we compare to other methods which are based on such assumptions, and we outperform them in our experiments.
With other data sets, potentially showing other characteristics (other tasks, other languages, other labeling functions), the performance could be different.
In particular, since the weakly supervised method performs almost as well as the supervised model, there is a need for tasks and datasets that are more challenging.

Another limitation is that currently our model is only implemented for classification, not for sequence tagging.
This adaptation would be possible, but is not trivial since sequential dependencies, unlabeled tokens, search, etc. would need to be carefully handled. 
We leave these extensions for future work.

\section*{Acknowledgements}

This research was funded by the WWTF through the project ”Knowledge-infused Deep Learning for Natural Language Processing” (WWTF Vienna Research Group VRG19-008).% and by the Deutsche Forschungsgemeinschaft (DFG, German Research Foundation).

%\subsection{Footnotes}

%Footnotes are inserted with the \verb|\footnote| command.\footnote{This is a footnote.}


\bibliography{latent_splitting}

\appendix

\section{Experimental Description}
\label{appendix:experimental_description}

As mentioned in the main paper, the code, setup and hyperparameters for the baselines are taken from the Wrench benchmark and are described in the corresponding paper \citep{zhang2021wrench}.
For the Gold+RoBERTa model there was no result available, as there are no ground truth labels for train set. Thus we assume an $F_1$-score of 45 for the computation of the average as it is better than the best performing model.

In the following paragraphs, the hyperparameters for the baselines we trained ourselves are briefly presented. 

\textbf{Weasel} \citep{cachay2021endtoend}. We experiment with hyperparameters similar to the original paper because there no transformer encoder "roberta-base" is used. We use AdamW \citep{loshchilov2018decoupled} and optimize hyperparameters for temperature in $\{0.33, 1.0\}$, dropout in $\{0.1, 0.2\}$,  weight decay in $\{$1e-4, 7e-7$\}$ and learning rate in $\{$1e-4, 1e-5,2e-5$\}$.
Unfortunately we were unable to obtain reasonable performance on the skewed datasets SMS and Spouse. 
The original paper does not use pre-trained language models, thus no direct comparison is possible. 
We decided to take the same encoder for all models.
%Additionally, we provide hyperparameters for Weasel and KnowMan
%Given that the benchmark paper might be updated, we repeat them here:

\textbf{KnowMAN} \citep{marz-etal-2021-knowman}. The values are set similar to the original paper. Batch size is 16, hidden size 700, dropout is 0.4 and trained is up to 5 epochs.
The classifier(C) and the discriminator(D) use Adam \cite{kingma2014} with a learning rate of 1e-4 and the feature extractor(F) uses AdamW \citep{loshchilov2018decoupled} with the same learning rate. For all three, \textit{num\_layer} is set to 1.

\section{Ablations}
\label{appendix:ablation}

\begin{table*}[ht]
    \centering
    \small
\begin{tabular}{lrrrrrrrrr}
\toprule
          &  IMDb &  Yelp &  Youtube &   SMS &  AGNews &  Trec & Spouse &  SemEval &      Avg. \\
\midrule
    Full model & 83.57 & 91.32 &    97.50 & 95.45 &   85.47 & 81.25 &   43.20 &    87.33 & 83.14 \\
    $-$ Weight decay & 83.57 & 91.32 &    97.50 & 95.45 &   85.47 & 79.03 &   42.55 &    86.99 & 82.73 \\
          $-$  $L_2$ reg. & 83.25 & 91.14 &    97.08 & 94.57 &   85.47 & 79.03 &   43.20 &    86.99 & 82.59 \\
  $-$ Unlabeled data & 82.13 & 88.34 &    97.50 & 95.45 &   85.32 & 81.25 &   39.89 &    86.99 & 82.11 \\
 $-$ Noise injection & 82.13 & 88.34 &    97.50 & 95.45 &   85.46 & 78.43 &   39.89 &    86.99 & 81.77 \\
     Basic model & 82.13 & 86.89 &    97.08 & 93.85 &   85.14 & 74.80 &   39.89 &    86.99 & 80.85 \\
\bottomrule
\end{tabular}
\caption{Extension of the ablation in Table \ref{tab:ablation_small}, showing all datasets.}
\label{tab:ablation_full}
\end{table*}

In addition to the summarized ablation in section \ref{exp_ablations},
Table \ref{tab:ablation_full} shows the impact of the ablations on task level. We observe that in general, the results for each task agree with the average.

%TODO: more data
%von Luisa: an könnte schon sowas wie nen coverage pro LF oder so haben, das wäre spannend. Aber manche datensets haben so viele LFS, da macht das ja keinen Sinn. Du könntest höchstens den Unterschied dann angeben von Max und Min coverage, aber das wird sicher auch verwirrend, muss dann extra erklärt werden und kostet so Platz

\section{Labeling Function Memorization}
\label{appendix:lf_memorization}

\begin{figure*}[h]
\centering
\begin{subfigure}{.31\textwidth}
  \centering
  \includegraphics[width=\linewidth]{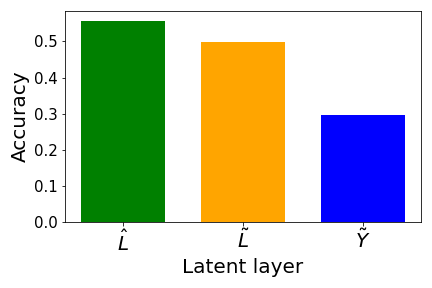}
  \label{fig:sub1}
\end{subfigure}%
$ $ $ $ $ $ $ $
\begin{subfigure}{.31\textwidth}
  \centering
  \includegraphics[width=\linewidth]{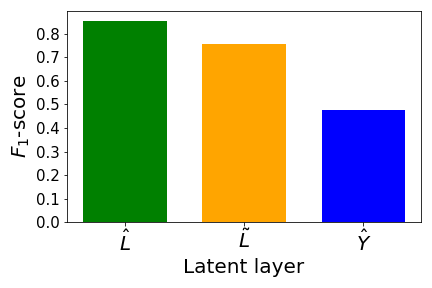}
  \label{fig:sub2}
\end{subfigure}
$ $ $ $ $ $ $ $
\begin{subfigure}{.31\textwidth}
  \centering
  \includegraphics[width=\linewidth]{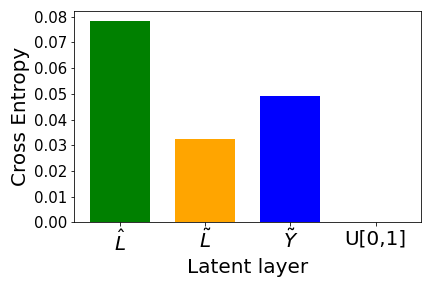}
  \label{fig:sub3}
\end{subfigure}
\caption{This is extension to Figure \ref{fig:lf_memorization}, describing the same metrics. But here we compute the the absolute difference between the metrics computed on the train set and the test set.
}
\label{fig:lf_memorization_train_test_diff}
\end{figure*}

The detailed numbers corresponding to the evaluation of section \ref{sec:lf_memorization} are given in Table \ref{tab:lf_preds_test_full}. 
The absolute difference between train and test split is shown in Figure \ref{fig:lf_memorization_train_test_diff}. The differences are low for accuracy, $F_1$-score and cross-entropy, thus we conclude that there is no or rather little overfitting towards the train set.

\begin{table*}[h]
    \centering
\begin{tabular}{lrrrrrrrrrc}
\toprule
 Dataset &  $\Tilde{L}$ Acc. &   $\hat{L}$ Acc. &  $\Tilde{Y}$ Acc. &  $\Tilde{L}$ $F_1$ &   $\hat{L}$ $F_1$ &  $\Tilde{Y}$ $F_1$ &  $\Tilde{L}$ CE &   $\hat{L}$ CE &  $\Tilde{Y}$ CE  &  $U[0, 1]$ CE \\
\midrule
    IMDb &       90.21 & 90.24 &  86.88 &      89.41 & 89.58 & 80.77 &       1.42 & 1.41 &  2.11 &    2.20 \\
    Yelp &       91.44 & 91.46 &  89.94 &      89.46 & 89.51 & 85.18 &       2.19 & 2.13 &  2.58 &    2.71 \\
 Youtube &       82.80 & 83.00 &  81.60 &      76.32 & 77.03 & 73.33 &       1.72 & 1.70 &  2.16 &    2.30 \\
     SMS &       96.41 & 95.65 &  99.28 &      97.59 & 97.19 & 98.92 &       4.26 & 5.31 &  4.76 &    4.29 \\
  AGNews &       91.05 & 91.54 &  89.48 &      89.06 & 90.45 & 84.50 &       1.72 & 1.66 &  1.92 &    2.20 \\
    Trec &       99.42 & 99.23 &  95.49 &      99.42 & 99.24 & 95.90 &       1.11 & 1.05 &  3.59 &    4.22 \\
  Spouse &       95.98 & 96.32 &  95.81 &      94.64 & 95.89 & 93.76 &       2.03 & 2.01 &  2.14 &    2.20 \\
 SemEval &       99.47 & 98.45 &  92.70 &      99.52 & 98.81 & 95.55 &       1.55 & 1.21 &  3.36 &    5.10 \\
 Avg. &       93.35 & 93.24 &  91.40 &      91.93 & 92.21 & 88.49 &       2.00 & 2.06 &  2.83 &    3.15 \\
\bottomrule
\end{tabular}
    \caption{    
Extension of Figure \ref{fig:lf_memorization}, showing all numbers explicitly. 
    The presented metrics are accuracy, macro $F_1$-score and cross entropy (CE) between true LF matrix $L$ (see section \ref{sec:method}) and the rows which are the full model $f_{\hat{L}}$, the latent layers $f_{\Tilde{L}}, f_{\Tilde{Y}}$ and the uniform distribution $\mathcal{U}[0,1]$.
    }
    \label{tab:lf_preds_test_full}
\end{table*}

\section{Impact of number of matches}
\label{appendix:impact_matches}

Table \ref{tab:lf_preds_test_full} provides detailed numbers for Figure \ref{fig:performance_by_matches}.
Since Figure \ref{fig:performance_by_matches} does not reflect how many samples correspond to a certain number of LF matches, detailed counts are added in Table \ref{tab:performance_by_matches}. 
This is an important addition as a performance measurement on a small number of samples is not indicative of the true performance. 
Luckily, usually the number of matching samples per number of matching labeling functions is distributed rather well.

\begin{table*}[h]
\centering
\begin{adjustbox}{width=0.4\textwidth}
\small
\parbox{.45\linewidth}{
\begin{tabular}{l c c c c c c}
\toprule
  &     0 &      1 &     2 &      3 &     4 &    5 \\
\midrule
    IMDb & 81.58 &  84.35 & 82.80 &   0.00 &  0.00 &  0.0 \\
    Yelp & 91.00 &  91.25 & 91.48 &  92.84 & 89.93 & 62.5 \\
 Youtube & 88.89 & 100.00 & 98.84 &  95.56 & 87.50 &  0.0 \\
     SMS & 95.12 &  97.44 & 88.89 & 100.00 &  0.00 &  0.0 \\
  AGNews & 77.48 &  88.38 & 90.81 &  87.50 & 87.50 &  0.0 \\
    Trec & 94.74 &  81.40 & 73.81 &  85.71 & 85.71 &  0.0 \\
  Spouse & 22.06 &  49.51 & 62.77 &  85.71 &  0.00 &  0.0 \\
 SemEval &  0.00 &  85.09 &  0.00 &   0.00 &  0.00 &  0.0 \\
\bottomrule
\end{tabular}
\caption{Detailed numbers corresponding to Figure \ref{fig:performance_by_matches}. The columns describe the number of matches a sample has, the cells the performance, i.e. accuracy or $F_1$-score.}
}
\end{adjustbox}
\hfill
\begin{adjustbox}{width=0.5\textwidth}
\centering
\small
\parbox{.55\linewidth}{
\begin{tabular}{l c c c c c c c}
\toprule
 \#matches &  total &     0 &     1 &     2 &    3 &    4 &   5 \\
\midrule
    IMDb &   2953 &   304 &  1521 &   593 &    0 &    0 &   0 \\
    Yelp &   5732 &   611 &  1463 &  1092 &  475 &  139 &  16 \\
 Youtube &    460 &    18 &    86 &    86 &   45 &    8 &   0 \\
     SMS &    263 &   302 &   148 &    37 &   12 &    0 &   0 \\
  AGNews &  11367 &  3699 &  5622 &  2318 &  336 &   24 &   0 \\
    Trec &    788 &    19 &   301 &    84 &   70 &   21 &   0 \\
  Spouse &   1019 &  1951 &   519 &   196 &   33 &    1 &   0 \\
 SemEval &   764.0 &     0 &   436 &    0.0 &   0.0 &   0.0 &  0.0 \\
\bottomrule
\end{tabular}

\caption{It is shown how many samples there are in total and split up in number of matches per sample.
Numbers are computed on the test set.}
\label{tab:performance_by_matches}
}
\end{adjustbox}
\end{table*}

\end{document}